\documentclass{article}
\usepackage{ijcai24}

\usepackage{times}
\usepackage{soul}
\usepackage{url}
\usepackage[hidelinks]{hyperref}
\usepackage[utf8]{inputenc}
\usepackage[small]{caption}
\usepackage{graphicx}
\usepackage{amsmath}
\usepackage{amsthm}
\usepackage{booktabs}
\usepackage{algorithm}
\usepackage{algorithmic}
\usepackage[switch]{lineno}
\usepackage{amssymb}
\usepackage{booktabs,makecell, multirow, tabularx}
\usepackage{subcaption}
\usepackage{amssymb}
\usepackage{dsfont}

\title{Effective High-order Graph Representation Learning for Credit Card \\Fraud Detection}
\author{
Yao Zou$^{1}$
\and
Dawei Cheng$^{1,2,3}$\footnote{Corresponding Author.}\\
\affiliations
$^1$Department of Computer Science and Technology, Tongji University, Shanghai, China\\
$^2$Key Laboratory of Artificial Intelligence, Ministry of Education, Shanghai, China\\
$^3$Shanghai Artificial Intelligence Laboratory, Shanghai, China\\
\emails
\{ski\_zy, dcheng\}@tongji.edu.cn
}

\begin{document}

\maketitle

\begin{abstract}
Credit card fraud imposes significant costs on both cardholders and issuing banks. Fraudsters often disguise their crimes, such as using legitimate transactions through several benign users to bypass anti-fraud detection. Existing graph neural network (GNN) models struggle with learning features of camouflaged, indirect multi-hop transactions due to their inherent over-smoothing issues in deep multi-layer aggregation, presenting a major challenge in detecting disguised relationships. Therefore, in this paper, we propose a novel High-order Graph Representation Learning model (HOGRL) to avoid incorporating excessive noise during the multi-layer aggregation process. In particular, HOGRL learns different orders of \emph{pure} representations directly from high-order transaction graphs. We realize this goal by effectively constructing high-order transaction graphs first and then learning the \emph{pure} representations of each order so that the model could identify fraudsters' multi-hop indirect transactions via multi-layer \emph{pure} feature learning. In addition, we introduce a mixture-of-expert attention mechanism to automatically determine the importance of different orders for jointly optimizing fraud detection performance. We conduct extensive experiments in both the open source and real-world datasets, the result demonstrates the significant improvements of our proposed HOGRL compared with state-of-the-art fraud detection baselines. HOGRL's superior performance also proves its effectiveness in addressing high-order fraud camouflage criminals.
\end{abstract}

\section{Introduction}
Credit card fraud significantly harms the financial health of individuals and businesses, has a major impact on the wider economy, undermines trust in the financial system, and disrupts the legal environment of society. Credit card fraud, typically conducted through credit or debit cards, refers to the unauthorized use of funds in a transaction \cite{BHATTACHARYYA2011602}. According to the Nilson Report, card fraud losses for issuers, merchants, and acquirers globally are projected to total $397.40$ billion over the next decade \cite{report1}. Obviously, credit card fraud has inflicted substantial economic losses, and effective credit card fraud detection is crucial for maintaining financial health and achieving the goals of Decent Work and Economic Growth.

\begin{figure}
    \centering
    \includegraphics[width=\linewidth]{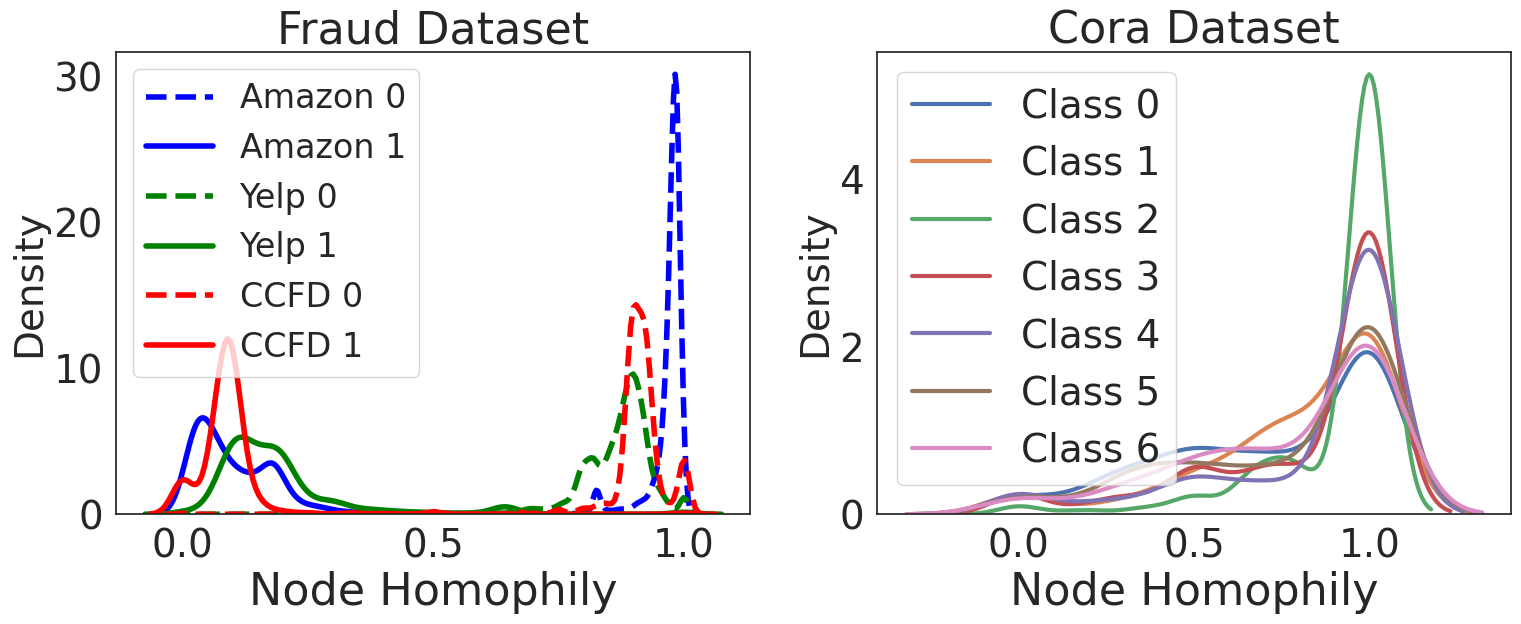}
    \caption{Node homophily distribution across datasets. Node homophily indicates the proportion of neighbors with the same label.}
    \label{fig:homo}
\end{figure}

Many models have been extensively researched and analyzed to address fraudulent transactions, ranging from rules-based approaches \cite{sanchez2009association,Seeja2014FraudMinerAN} to machine learning methods \cite{burrell2016machine}. Later, deep learning models have been developed to uncover latent fraud patterns \cite{fiore2019using,ma2023fighting}. However, these methods treat each fraud action as isolated, lacking the capability to identify more sophisticated and covert transactions. Recently, graph neural networks (GNNs) have been employed for credit card fraud detection \cite{wang2019semi,cheng2020graph,zhang2024pre} and achieve remarkable success as GNNs can more accurately infer the fraud probability by capturing patterns within relational graphs.

However, current fraudsters often use multiple benign entities as intermediaries for indirect transactions to disguise themselves and avoid being identified as part of a fraud ring \cite{liu2021pick}. This disguise behavior means that a fraudster's direct neighbors might predominantly be benign entities, challenging the assumption in GNNs that entities of similar categories connect more frequently, which could compromise the model's effectiveness \cite{platonov2024characterizing}. As depicted in Figure \ref{fig:homo}, we analyzed the proportion of nodes connecting with neighbors of the same label (referred to as node homophily) in three fraud datasets and one general dataset (Cora). Fraudulent nodes are mainly distributed around 0, exhibiting low homophily, providing evidence for their deceptive behavior, while benign nodes are primarily distributed around 1, showing high homophily. Furthermore, within the Cora dataset, different categories often display higher levels of homophily, whereas in the fraud dataset, categories of varying types exhibit completely different homophily distributions. Some studies have acknowledged the challenges posed by the deceptive behavior of fraudsters \cite{liu2020alleviating,shi2022h2,meng2023generative}.

However, cunning fraudsters may engage in these indirect transactions through four or even more unsuspecting benign entities. The aforementioned methods require deeper network layers to identify such multi-hop indirect transactions, but increasing depth leads to the phenomenon of feature over-smoothing \cite{li2018deeper,chien2021adaptive}. Although some high-order GNNs \cite{he2021bernnet,wang2022powerful,bo2021beyond} have developed to alleviate the oversmoothing issue, they mostly rely on mixed-order propagation. This entails the mixing of high-order and low-order information before propagation to the central node \cite{feng2020graph}. However, in disguise scenarios, these approaches \cite{wang2021tree} face significant challenges as the propagation characteristics inevitably lead to the blending of high-order information with low-order noise, consequently producing non-discriminative node representations.

Therefore, to address the issue of disguised fraud involving multi-hop indirect transactions, we propose a novel High-order Graph Representation Learning model (HOGRL), aiming to avoid introducing excessive noise during the multi-layer aggregation process. Specifically, HOGRL directly learns distinct orders of pure representations from high-order transaction graphs. We achieve this by effectively decoupling neighbor nodes across different hierarchical levels to construct high-order transaction graphs. Subsequently, we learn the pure representations of each high-order graph, allowing the model to recognize multi-hop indirect transactions by means of multi-layer pure feature learning for identifying concealed fraudsters. Additionally, we introduce mixture-of-expert attention mechanism to automatically determine the significance of different orders, thus jointly optimizing fraud detection performance. Considering the potential loss of original structural information in the constructed multi-layer high-order transaction graphs, HOGRL combines embeddings from the original graph and multi-layer high-order transaction graphs into the final node representation. Extensive experiments conducted on a real credit card dataset and two public fraud datasets demonstrate the superior performance of HOGRL compared with state-of-the-art baselines. Contributions of our work are summarized as follows:
\begin{itemize}
    \item  We propose a high-order graph representation learning model to address the issue of disguised fraudsters engaging in multi-hop indirect transactions.
    \item We effectively construct high-order transaction graphs and directly learn distinct orders of pure representations from them. Additionally, we introduce mixture-of-expert attention mechanism to automatically determine the significance of different orders, thereby jointly optimizing the learning process.
    \item We conduct extensive experiments to compare our method with state-of-the-art baselines on both public and real-world datasets. The results show the superiority of our proposed HOGRL on fraud detection.
\end{itemize}
\section{Related Works}
\subsection{GNN-based Fraud Detection}
Several machine learning techniques have been proposed in the literature to address the problem of fraud detection \cite{panigrahi2009credit,Fu2016CreditCF,niu2020iconviz}. Recently, techniques based on GNNs have been introduced for fraud detection \cite{cheng2020graph,xiang2023semi}.
However, current fraudsters employ sophisticated disguises to evade detection. Some studies have noticed similar challenges. CAREGNN \cite{dou2020enhancing} employs label-aware similarity measurement and reinforcement learning modules to select more informative neighbors. PCGNN \cite{liu2021pick} employs balanced sampling and selective neighbor aggregation for node representation. Some studies attribute it to the heterogeneity of the graph \cite{liu2018heterogeneous,cheng2023anti}. For instance, H$^{2}$-FDetector \cite{shi2022h2} leverages homophilic and heterophilic interactions along with a specialized aggregation strategy and category prototypes to enhance detection effectiveness. However, the mentioned approaches struggle with oversmoothing in identifying multi-hop indirect disguised transactions. In contrast, HOGRL learns pure representations directly from high-order transaction graphs, facilitating the recognition of multi-hop indirect disguised transactions through layered configurations.

\subsection{High-order Graph Neural Networks}
Recently, scholars have started to tackle the problem of shallow layers in GNNs. \cite{li2019deepgcns} adopted the ResNet \cite{he2016deep} concept  from image processing, enabling the construction of deep network structures with dozens of layers in GNNs. GPRGNN \cite{chien2021adaptive} introduces a novel Generalized PageRank architecture, assigning learnable weights to enable deep learning capabilities in the model. There are also some studies \cite{he2021bernnet,wang2022powerful} that attempt alternative methods of learning weights. MHGNN \cite{xue2020multi} expands the receptive field by utilizing multi-hop node information, enabling the capture of nodes within multiple hops in a single layer. FAGCN \cite{bo2021beyond} introduces an adaptive graph convolutional network with a self-gating mechanism to simultaneously capture both low-order and high-order information. AdaGNN \cite{dong2021adagnn} incorporates a trainable filter design that spans across multiple layers to capture the varying importance of different frequency components for node representation learning. Although these methods theoretically detect fraud through high-order information, in practical applications, due to mixed-order propagation, integrating high-order multi-hop information can blend with low-order noise. HOGRL, by decoupling neighbors at different orders to construct high-order transaction graphs and directly learning pure representations at different orders, avoids the mixture of noise during the aggregation process.
\begin{figure*}
    \centering
    \includegraphics[width=\linewidth]{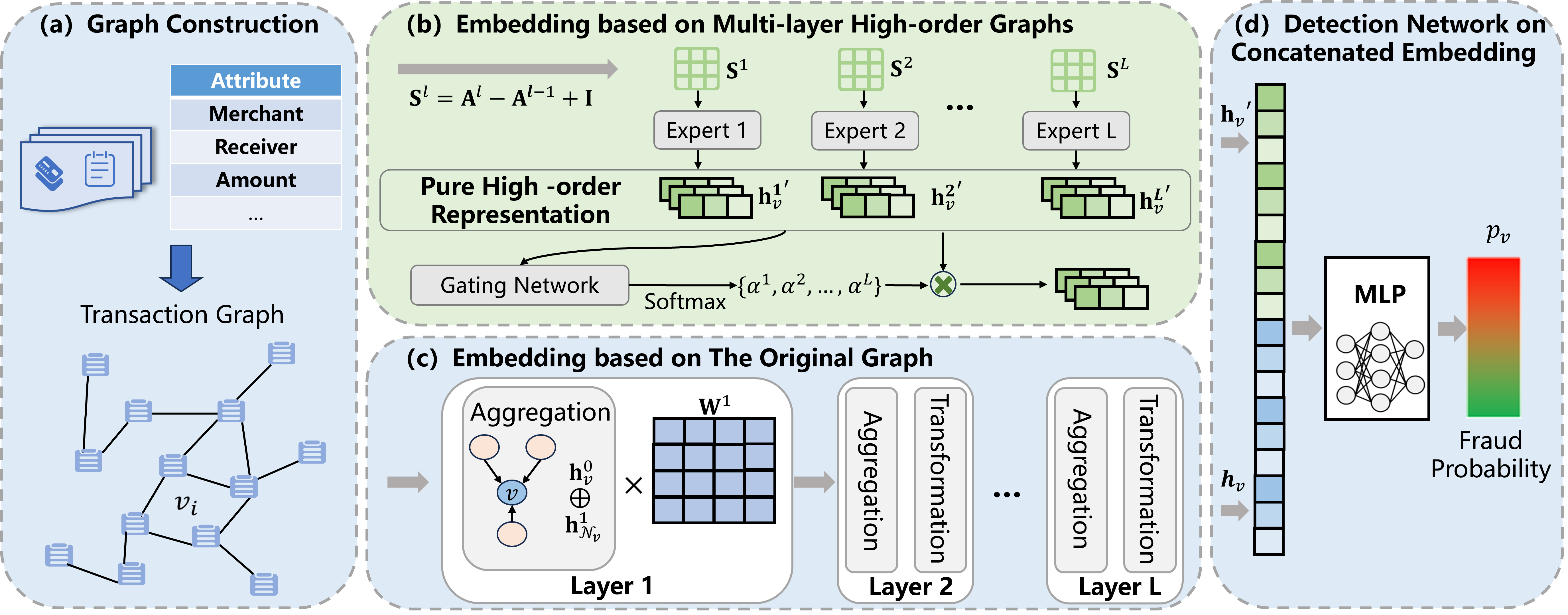}
    \caption{ The illustration of the proposed HOGRL model architecture. It contains four components: (a) Graph construction based on original transactions. We treat each layer of the GNN network as an expert network and dynamically allocate weights using a mixture-of-expert attention mechanism. (b) Embedding generation based on multi-layer high-order graphs. (c) Graph neural network embedding generation based on the original graph. (d) Detection network based on concatenated embeddings and joint optimization.}
    \label{fig:model}
\end{figure*}
\section{Methodology}
As shown in Figure \ref{fig:model}, our model is primarily divided into four parts: the construction of the transaction graph, the generation of node embeddings based on both multi-layered high-order graphs and the original graph, and the detection network. In this section, we first introduce relevant definitions and the construction of the credit card fraud transaction graph. Following that, we introduce how to construct high-order transaction graphs and obtain pure high-order representations from them, as well as generate embeddings based on the original graph. Finally, we introduce the detection network and optimization strategy.
\subsection{Preliminaries}
\noindent\textbf{Node homophily.} The homophily of node \cite{pei2020geom} \(v\) represents the proportion of its neighbors that have the same label as \(v\), which can be expressed as:
\begin{equation}
    \begin{aligned}
        \mathcal{H}(v)=\frac{\left | \left \{ y_{u}==y_{v},u \in \mathcal{N}_{v} \right \} \right | }{\left | \mathcal{N}_{v} \right | } .
    \end{aligned}
\end{equation}

In credit card fraud detection, we define the credit card transaction graph as $\mathcal{G}=(\mathcal{V},\mathcal{E})$, where $\mathcal{V}=\left(v_{1}, v_{2}, \ldots, v_n\right)$ denotes a set of credit card transactions (i.e, $n=|\mathcal{V}|$, we call it node), and $\mathcal{E} \subset \mathcal{V} \times \mathcal{V}$ represents the set of $m$ edges (i.e, $m=|\mathcal{E}|$) between transactions in $\mathcal{V}$ with $e_{u,v}$ denoting that the transaction $u$ and the transaction $v$ have the same merchant or receiver, and $\mathbf{X} \in \mathbb{R}^{n \times d}$ denotes the feature matrix, where each row $\boldsymbol{x}_{i} \in \mathbb{R}^{d}$ represents the feature of vector of the node $v_{i}$ and $d$ is the dimension of node features. We define $\mathcal{Y} = \{y_{1}, \ldots, y_{n}\}$ as the set of fraud labels, where $y_{i} \in \{0, 1\}$ with 0 representing normal and 1 representing fraud. The topological information of the $\mathcal{G}$ is described by the adjacency matrix $\mathbf{A} \in \mathbb{R}^{n \times n}$, where $\mathbf{A}_{u,v} = 1$ if an edge exists between the node $u$ and the node $v$. $\mathcal{N}_{v}$ is the neighborhood node set of the center node $v$, which is given by $\mathcal{N}_{v}=\left\{u \mid e_{u,v} \in \mathcal{E}\right\}$. We extend this definition by using $\mathcal{N}_{v}^{l}$ to denote the $l$-th layer neighborhood nodes of the center node $v$, which includes all nodes that can be reached from the center node $v$ in exactly $l$ hops. For each credit card record, we aim to infer the possibility of whether it is a fraud event, and our task can be formulated a node classification.
\subsection{High-Order Graphs Generation}
The fundamental assumption of GNNs is that leveraging neighborhood information through feature propagation and aggregation can enhance the predictive performance of the central node \cite{xu2018representation}. This assumption is based on the premise that connected nodes tend to share similar features and usually belong to the same category \cite{pei2020geom,paszke2019pytorch}. However, in scenarios involving camouflage, the fraudulent nodes' indirect transaction disguises may result in low-order neighbors of fraudsters being multiple benign nodes, which contradicts the premise. This implies the need to introduce higher-order fraudulent information for identifying disguised fraudsters.

Nevertheless, existing high-order GNNs mostly rely on mixed-order propagation, where while introducing higher-order information, all layers of information are mixed and propagated to the central node. This can lead to contamination of high-order effective information by low-order noise information. An intuitive idea is to allow high-order information to be directly conveyed to the central node without mixing, ensuring purer higher-order information. Therefore, we propose decoupling neighbors at different orders to construct high-order transaction graphs. The high-order transaction graph for the $l$-th layer only includes neighboring nodes that can be reached within at least $l$ hops. The adjacency matrix of the high-order transaction graph for the $l$-th layer can be represented as follows:
\begin{equation}
\begin{aligned}
    \mathbf{S}^{l}=\mathbf{A}^{l}-\mathbf{A}^{l-1}+\mathbf{I},
    \label{eq:decomposion}
\end{aligned}
\end{equation}
where $\mathbf{A}^0=\mathbf{I}$  is the identity matrix.
$\mathcal{N}_{v}(\mathbf{S}^{l})$ represents the set of neighboring nodes of node $v$ under the adjacency matrix $\mathbf{S}^{l}$, which can be represented as:
\begin{equation}
    \begin{aligned}
        \mathcal{N}_{v}\left(\mathbf{S}^{l}\right)=\left(\mathcal{N}_{v}^{l} \backslash\left(\mathcal{N}_{v}^{l} \cap \mathcal{N}_{v}^{l-1}\right)\right) \cup\{v\}
    \end{aligned}
\end{equation}
\subsection{Graph Representation Learning}
For the $l$-th layer high-order transaction graph, the aggregation process can be represented as:
\begin{equation}
\begin{aligned}
    {\mathbf{h}^{l}}^{'}=\text{ReLU}\left ( \mathbf{S}^{l}\cdot\mathbf{X}\cdot{\mathbf{W}^{l}}^{'} \right ) ,
    \label{eq:pure_order}
\end{aligned}
\end{equation}
where ${\mathbf{W}^{l}}^{'}$ is the parameter matrix of the $l$-th layer high-order transaction graph. Due to the varying contributions of each layer's high-order transaction graph to the final node embeddings, we employ a mixture-of-expert attention mechanism to automatically determine the importance of different layer high-order transaction graphs. Specifically, we treat each layer of the graph neural network (Eq(\ref{eq:pure_order})), as an individual expert network. The intermediate representations generated by these layers are considered the outputs of the expert networks. A gating network is then utilized to distribute weights across the outputs from each expert network. For the high-order transaction graph at the $l$-th layer, the weight allocated can be articulated as follows:
\begin{equation}
    \begin{aligned}
    f_{l}\left({\mathbf{h}^{l}}^{'}\right)=\boldsymbol{w}_{l}^{\text{T}}\cdot {\mathbf{h}^{l}}^{'}+\boldsymbol{b}_{l}
    \end{aligned}
\end{equation}
\begin{equation}
    \begin{aligned}
        \alpha^{l}=\frac{\exp \left(f_{l}\left({\mathbf{h}^{l}}^{'}\right)\right)}{\sum_{k}^{L} \exp \left(f_{k}\left({\mathbf{h}^{k}}^{'}\right)\right)},
    \end{aligned}
\end{equation}
where \(\boldsymbol{w}_l\) is the weight vector for the \(l\)-th expert from the gating network, and \(b_e\) is the bias term. It's important to note that while the design of the gating network's weights is similar to traditional attention mechanisms, its goal is to dynamically adjust the influence of each expert network's output on the final output, aligning with the central idea of a mixture-of-experts. Then, the embeddings generated based on multi-layer high-order graphs can be represented as:
\begin{equation}
\begin{aligned}
    \mathbf{h}^{'}=\sum_{l=1}^{L}  \alpha^{l}{\mathbf{h}^{l}}^{'}.
\end{aligned}
\end{equation}
The mixture-of-expert attention mechanism enables the model to adaptively select more informative hierarchical features, thereby improving the overall performance of the model. Then, we delve into the process of generating embeddings from the original graph. In the context of the original graph, we adopt the mean operator as the aggregator within the GNN, which is represented as follows:
\begin{equation}
\begin{aligned}
    \mathbf{h}_{v}^{l} = \text{ReLU}\left(\mathbf{W}^{l} \cdot \left(\mathbf{h}_{v}^{l-1}\oplus\mathbf{h}_{\mathcal{N}_{v}}^{l}\right)\right),
\end{aligned}
\end{equation}
\begin{equation}
\begin{aligned}
    \mathbf{h}_{\mathcal{N}_{v}}^{l} = \operatorname{MEAN}\left(\left\{\mathbf{h}_{u}^{l-1}, \forall u \in \mathcal{N}_{v}\right\}\right),
\end{aligned}
\end{equation}
where $\mathbf{h}_{v}^{0}= \boldsymbol{x}_{v}$, the $\mathbf{W}^{l}\in \mathbb{R}^{d_{l}\times d_{l-1}}$ is the $l$-th parameter matrix and $\oplus$ denotes the concat operation.
We combine the learned $\mathbf{h}_{v}^{L}$ (represented it as $\mathbf{h}_{v}$ for simplicity.) with $\mathbf{h}_{v}^{'}$ as the last representation:
\begin{equation}
    \begin{aligned}
    \mathbf{z}_{v} = \mathbf{h}_{v}+\gamma\mathbf{h}_{v}^{'},
    \end{aligned}
\end{equation}
where $\gamma$ is a hyperparameter that determines the weight of embeddings generated based on multi-layer high-order graphs. Integrating embeddings generated from the original graph with multi-layer high-order graphs is based on the following concept: Although the constructed high-order graphs can directly transmit high-order information to the central node, they lose the original multi-hop dependencies \cite{wang2021tree}. Specifically, the $l$-th layer high-order graph includes connections reached exactly within at least $l$ hops, neglecting intermediate nodal entities. By integrating embeddings derived from the original graph, it becomes feasible to preserve multi-hop dependencies. This strategy enhances information propagation efficiency and maintains critical pathway dependencies within the network's framework, thereby enriching the insight available for advanced network structure analysis and understanding.

For a multi-relational graph $\mathcal{G}=(\mathcal{V},\mathbf{E})$, where $\mathbf{E}=\left\{\mathcal{E}_{1}, \ldots, \mathcal{E}_{R}\right\}$ is the edge set of $R$ relations, we perform graph propagation separately for each relation and concatenate the embeddings. This can be represented as:
\begin{equation}
    \begin{aligned}
        \mathbf{z}_{v} = (\mathbf{z}_{v}^{(1)}\oplus\mathbf{z}_{v}^{(2)}\oplus\dots \oplus \mathbf{z}_{v}^{(R)}).
    \end{aligned}
\end{equation}
\subsection{Detection Network and Optimization}
In the downstream detection task, we utilize a multi-layer perceptron (MLP) as the detection network to infer the fraud probability as:
\begin{equation}
\begin{aligned}
    p_{v}=\mathbf{MLP}(\mathbf{z}_{v}).
\end{aligned}
\end{equation}

For the node classification task, we adopt the cross-entropy
loss function for optimization, which can be formulated as:
\begin{equation}
\begin{aligned}
    \mathcal{L}_{\mathrm{gnn} } = -\sum_{v \in \mathcal{V} }\left [ y_{v} \log{p_{v}}+\left ( 1-y_{v}\right ) \log{(1-p_{v})}   \right ],
\end{aligned}
\end{equation}
where $y_v \in \mathcal{Y}$ is the label of the node $v$. The proposed method can be optimized through the standard
stochastic gradient descent-based algorithms. In this paper,
we used the Adam optimizer \cite{Kingma2014AdamAM} to learn
the parameters. We set the initial learning rate to $5\times{10}^{-3}$ and the weight decay to $5\times{10}^{-5}$ by default.
\subsection{Complexity Analysis}
Compared to traditional GCNs, the additional computational burden in our approach primarily stems from generating node intermediate representations based on high-order transaction graphs. There is no need to calculate $A^{l}$. We can calculate $\mathbf{S}^{l}\cdot\mathbf{X}$ with right-to-left multiplication. The calculation process can be represented as follows:
\begin{equation}
\begin{aligned}
\mathbf{S}^{l} \cdot \mathbf{X} &= (\mathbf{A}^{l} - \mathbf{A}^{l-1} + \mathbf{I}) \cdot \mathbf{X} \\
&= \mathbf{A}^{l}\mathbf{X} - \mathbf{A}^{l-1}\mathbf{X} + \mathbf{X} \\
&= ( \underbrace{\mathbf{A}(\mathbf{A}(\ldots(\mathbf{A}}_{l} \cdot \mathbf{X})) ) + ( \underbrace{\mathbf{A}(\mathbf{A}(\ldots(\mathbf{A}}_{l-1} \cdot \mathbf{X})) ) + \mathbf{X}
\end{aligned}
\end{equation}
If we store $\mathbf{A}$ as a sparse matrix with $m$ non-zero entries, then the embeddings generated by the $l$-th order transaction graph require $O(l \times m \times d)$ computational time, where $d$ is the feature dimension of $\mathbf{X}$. Under the realistic assumptions that $l \ll m$ and $d \ll m$, running an $L$-order layer requires $O(Lm)$ computational time. This matches the computational complexity of the traditional GCN.
\section{Experiments}
\subsection{Experimental Settings}
\noindent\textbf{Datasets.} We have collected fraudulent transaction data from a major commercial bank, including real-world credit card transaction records, involving a total of $476,124$ different users. The ground truth labels are based on consumer reports, verified by financial domain experts. Transactions reported as fraud or confirmed by experts were marked as 1, while non-fraudulent transactions were marked as 0. We refer to this dataset as CCFD (Credit Card Fraud Detection Dataset). Besides, we also experimented on two public fraud detection
datasets. The YelpChi graph dataset \cite{rayana2015collective} contains a selection of hotel and restaurant reviews on Yelp. There are three edge types in the graph, including R-U-R (the reviews posted by the same user), R-S-R (the reviews under the same product with the same star rating),
and R-T-R (the reviews under the same product posted in the same month). The Amazon graph dataset \cite{mcauley2013amateurs} includes product reviews of
musical instruments. There are also three relations: U-P-U (users
reviewing at least one same product), U-S-U (users having
at least one same star rating within one week), and U-V-U
(users with top-5\% mutual review TF-IDF similarities). CCFD is a single-relation graph. Some basic statistics of three fraud datasets are shown in Table \ref{table:1}.
\begin{table}
\centering
\begin{tabular}{c c c c}
\hline
\textbf{Dataset} & \textbf{\#Node} & \textbf{Relations} & \textbf{\#Relations} \\ \hline
\multirow{3}{*}{YelpChi} & \multirow{3}{*}{45,954} & R-U-R & 49,315 \\
                         &                         & R-S-R & 3,402,743 \\
                         &                         & R-T-R & 573,616 \\ \hline
\multirow{3}{*}{Amazon}  & \multirow{3}{*}{11,944} & U-P-U & 175,608 \\
                         &                         & U-S-U & 3,566,479 \\
                         &                         & U-V-U & 1,036,737 \\ \hline
CCFD                     & 1,820,840               & -     & 31,619,440 \\ \hline
\end{tabular}
\caption{Statistics of three Datasets.}
\label{table:1}
\end{table}

\noindent\textbf{Compared Baselines.} We compare with several state-of-the-art GNN-based methods to verify the effectiveness of HOGRL: GCN \cite{kipf2016semi}, GAT \cite{veličković2018graph}, Graphsage \cite{hamilton2017inductive}, GPRGNN \cite{chien2021adaptive}, FAGCN \cite{bo2021beyond}, GraphConsis \cite{liu2020alleviating}, CARE-GNN \cite{dou2020enhancing}, PC-GNN \cite{liu2021pick}, H$^{2}$-FDetector \cite{shi2022h2}, GTAN \cite{xiang2023semi}, BWGNN \cite{tang2022rethinking}.

\noindent \textbf{Metrics and Implementation.} For class imbalance classification, the evaluation
metrics should have no bias to any class \cite{luque2019impact}. Therefore, We evaluate the experimental results on three fraud datasets by the area under the ROC curve (AUC), macro average of F1-macro score (F1-macro), and GMean (Geometric Mean).

For all baselines, if the original hyperparameters are provided, we use them. If not, the hyper-parameter search space is: learning rate
in \{0.01, 0.05, 0.001\}, dropout in \{0.3, 0.4, 0.5, 0.6\}, weight decay in
\{$10^{-3}$, $10^{-4}$, $10^{-5}$\} hidden dimension in \{16, 32, 64\}. For high-order GNNs, we explore the number of layers \{1, 2, ... , 9\}. For HOGRL, we set the batch size to 2048 for yelp and CCFD, 256 for amazon, the dropout ratio to 0.3, the embedding dimension to 64 ($\mathbf{h}_{v} $ and $ {\mathbf{h}_{v}}^{'}$), the number of layers to 7, and the maxinum number of epochs is set to 1000. The train, val and test are set to be 40\%, 40\%, 20\% respectively. We train and test on the validation set every 10 times, and select the model that performs best on the validation set to test after training ends. Our method is implemented using PyTorch 1.12.1 with CUDA 11.2 and Python 3.7 as the backend. The model is trained on a server with two 32GB NVIDIA Tesla V100 GPUs.
\subsection{Fraud Detection Performance}
\begin{table*}[!htbp]
\tabcolsep 3.5pt
\centering
\begin{tabular}{cc|ccccccccccccc}
\toprule
 \multicolumn{2}{c|}{\multirow{2}{*}{\textbf{Model}}} & \multicolumn{3}{c}{\textbf{YelpChi}} & &  \multicolumn{3}{c}{\textbf{Amazon}}  & &
    \multicolumn{3}{c}{\textbf{CCFD}}\\
    \cmidrule{3-5} \cmidrule{7-9} \cmidrule{11-13}
    & & F1-macro & AUC & GMean && F1-macro & AUC & GMean && F1-macro & AUC & GMean\\

    \midrule
\multirow{3}{*}{Traditional}
&GCN
& 0.5735 & 0.6128 & 0.5752 &
& 0.6438 & 0.8422 & 0.7793 &
& 0.4873 & 0.5236 & 0.5192
\\
&GAT
& 0.5019 & 0.6171 & 0.2072 &
& 0.5089 & 0.8503 & 0.2045 &
& 0.4958 & 0.5405 & 0.6232
\\
&GraphSage
& 0.6548 & 0.8351 & 0.7597 &
& 0.7849 & 0.9519 & 0.9084 &
& 0.5104 & 0.5444 & 0.5011
\\
\midrule
\multirow{2}{*}{High-order}
&FAGCN
& 0.6256 & 0.7583 & 0.6667 &
& 0.8719 & 0.9644 & 0.8898 &
& 0.6621 & 0.7444 & 0.6545
\\
&GPRGNN
& 0.6086 & 0.7503 & 0.6954 &
& 0.8023 & 0.9546 & 0.8825 &
& 0.5974 & 0.7389 & 0.6587
\\
\midrule
\multirow{5}{*}{\shortstack{Fraud \\ Detection}}
&GraphConsis
& 0.5673 & 0.6985 & 0.6182 &
& 0.7378 & 0.8836 & 0.7391 &
& 0.6506 & 0.6053 & 0.4626
\\
&CARE-GNN
& 0.6332 & 0.7619 & 0.6791 &
& 0.8946 & 0.9067 & 0.8962 &
& 0.5771 & 0.6623 & 0.5728
\\
&PC-GNN
& 0.6300 & 0.7987 & 0.7160 &
& 0.8999 & 0.9585 & 0.8995 &
& 0.6077 & 0.6795 & 0.5929
\\
&H$^{2}$-FDetector
& 0.6944 & 0.8877 & 0.8160 &
& 0.8470 & 0.9711 & 0.9223 &
& 0.6531 & 0.6739 & 0.6163
\\
&GTAN
& 0.7788 & 0.9141 & 0.8821 &
& 0.9213 & 0.9621 & 0.9081 &
& \textbf{0.6913} & 0.7218 & 0.6291
\\
&BWGNN
& 0.7891 & 0.9170 & 0.8791 &
& 0.9191 & 0.9759 & 0.9195 &
& 0.6856 & 0.7195 & 0.6193
\\
\midrule
\multirow{1}{*}{Ours}
&\textbf{HOGRL}
& \textbf{0.8595*} & \textbf{0.9808*} & \textbf{0.9361*} &
& \textbf{0.9198*}  &\textbf{0.9800*} & \textbf{0.9438*} &
& 0.6861 & \textbf{0.7590*} & \textbf{0.6784*}
\\ \bottomrule
\end{tabular}
\caption{Fraud detection performance on three datasets compared with popular benchmark methods.
}
\label{tab:2}
\end{table*}
We repeat the experiments ten times for each method and show the average performance in Table \ref{tab:2}. $*$ denotes that the improvements are statistically signifcant for $p<$ 0.01 according to the paired $t$-test.

The first three rows of Table 2 report the results of some classic graph-based methods, including GCN, GAT, GraphSage. It is clear that the results of GCN and GAT not satisfactory, showing the limitation of traditional GNNs-based model in addressing the complex fraud patterns. GraphSage improves performance, attributed to its suitability for large graphs. FAGCN and GPRGNN can capture higher-order information, thus performing well. They even outperform all graph-based fraud detection methods on the CCFD dataset, further highlighting the importance of higher-order information in identifying disguised fraudsters. The graph-based fraud detection methods (GraphConsis, CARE-GNN, PC-GNN, H$^{2}$-FDetector) focus on the deceptive behaviors of fraudsters, but their models still perform lower than HOGRL due to their shallow model limitations. The comparison with the semi-supervised model GTAN is detailed in Section 4.4.

BWGNN excels by employing customized spectral filters to capture effective information of fraudsters, making it the state-of-the-art method for graph-based anomaly detection, while HOGRL outperforms it significantly. On the YelpChi dataset, compared to BWGNN, HOGRL achieves improvements of 8.9\% in F1-macro, 6.9\% in AUC, and 6.4\% in GMean. It's worth noting that popular fraud detection models perform poorly on CCFD, mainly due to the presence of complex fraudulent techniques in real-world scenarios, affecting fraud models' performance. However, our proposed HOGRL outperforms other models on this dataset. We achieves the best results with an improvement of 5.4\% in AUC compared to the BWGNN, and a 9.5\% improvement in GMean score. Compared to high-order GNNs, HOGRL shows an improvement of 1.9\% in AUC score, 3.6\% in F1 Score, and 2.9\% in GMean score at least.
\subsection{Ablation Study}
\begin{figure}
    \centering
    \begin{subfigure}[b]{0.49\linewidth}
        \includegraphics[width=\textwidth]{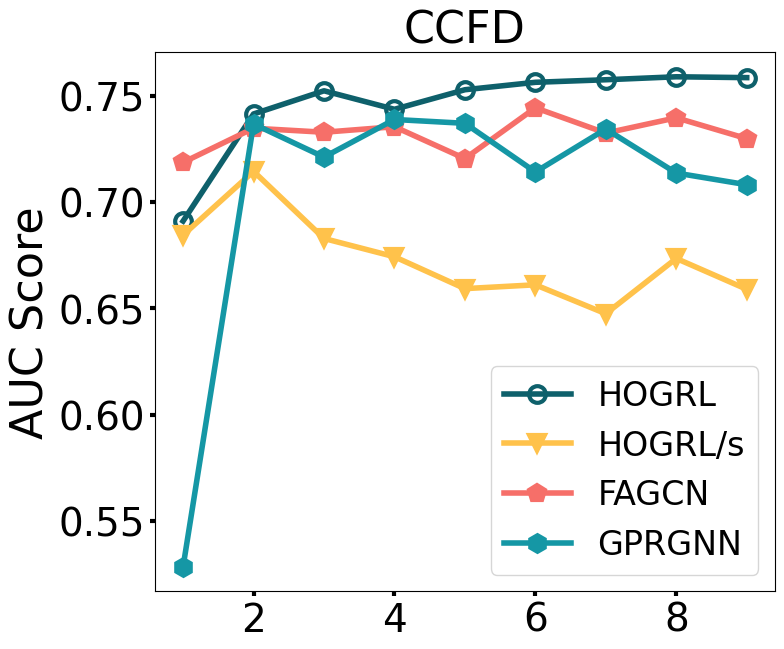}
    \label{fig:yelp_auc}
    \vspace{-12pt}
    \end{subfigure}
    \begin{subfigure}[b]{0.49\linewidth}
        \includegraphics[width=\textwidth]{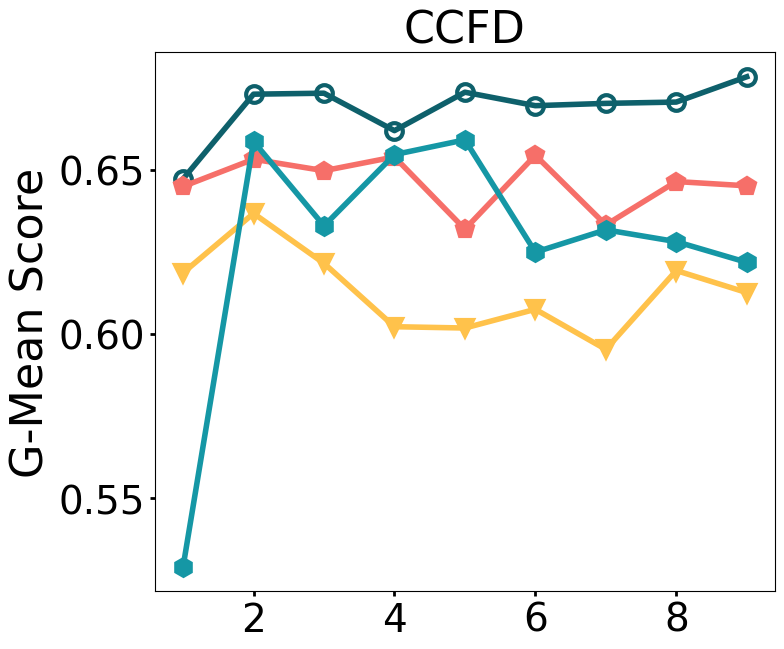}
    \label{fig:yelp_gmean}
    \vspace{-12pt}
    \end{subfigure}
    \begin{subfigure}[b]{0.49\linewidth}
        \includegraphics[width=\textwidth]{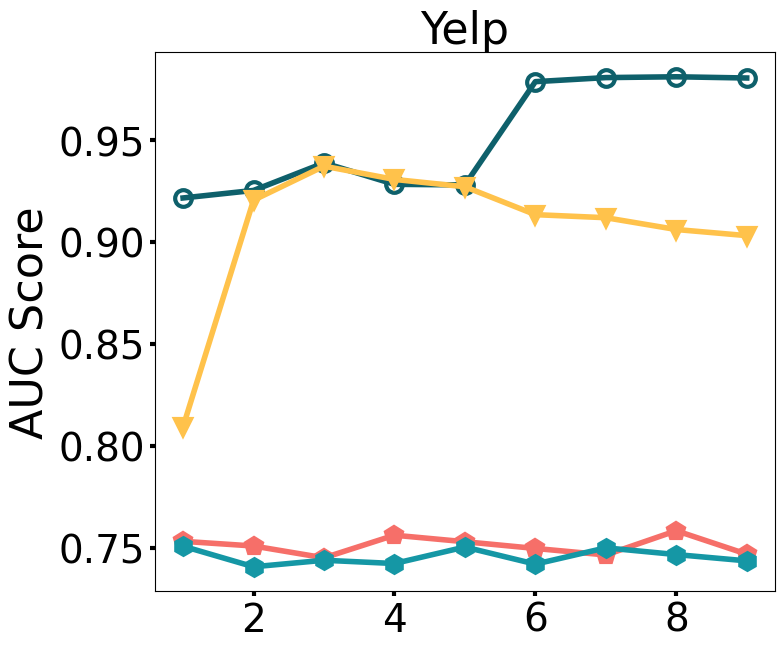}
    \label{fig:strad_auc}
    \vspace{-5pt}
    \end{subfigure}
    \begin{subfigure}[b]{0.49\linewidth}
        \includegraphics[width=\textwidth]{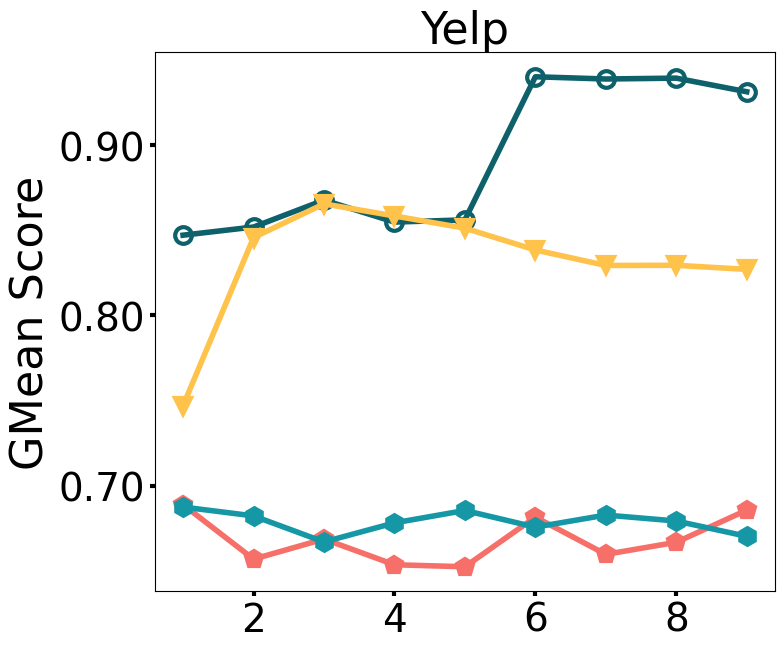}
    \label{fig:strad_gmean}
    \vspace{-5pt}
    \end{subfigure}
    \vspace{-10pt}
    \caption{Results of models with different layers.}
    \label{fig:multi_layer}
    \vspace{-10pt}
\end{figure}

To validate the effectiveness of constructing pure representations using high-order graphs, we designed HOGRL/s, a model that generates embeddings solely based on the original graph. We chose FAGCN and GPRGNN as controls to explore HOGRL's ability to capture high-order information.
We visualize the performance of each model using
layers from up to 1 to up to 9 in Figure \ref{fig:multi_layer}. It is evident that HOGRL/s achieves its highest performance on the Yelp dataset when $L$=3, and on the CCFD dataset when $L$=2. This pattern is consistent with most GNNs, where an increase in the number of layers beyond a certain point leads to a decline in performance. The reason for this decline is attributed to the over-smoothing caused by the coupling of multiple features. However, HOGRL directly generates pure higher-order representations from high-order graphs, alleviating the oversmoothing caused by feature coupling. It is observable that on the Yelp dataset, as the number of layers increases to 6, HOGRL significantly outperforms HOGRL/s in both AUC and Gmean metrics. On the CCFD dataset, a clear difference between the two is evident when the number of layers reaches 3. As the number of layers continues to increase, the performance of HOGRL/s steadily declines, whereas the performance of HOGRL remains stable and peaks at $L$=9. We set $L$ to 7 due to the computational complexity.

In comparison with higher-order GNNs, it is observed that the performance of FAGCN and GPRGNN on the Yelp and CCFD datasets exhibits significant fluctuations with an increase in the number of network layers, highlighting their lack of stability. Notably, on the Yelp dataset, HOGRL significantly outperforms FAGCN and GPRGNN in terms of AUC and GMean metrics. On the CCFD dataset, except for when the layer count L is 1, where HOGRL's AUC performance is slightly inferior to FAGCN, in all other cases it surpasses both FAGCN and GPRGNN. These results further emphasize the substantial advantage of HOGRL in capturing higher-order information, thereby better identifying disguised fraudsters.
\subsection{Parameter Sensitivity}
\begin{figure}
  \centering
  \begin{subfigure}[b]{0.49\linewidth}
    \includegraphics[width=\textwidth]{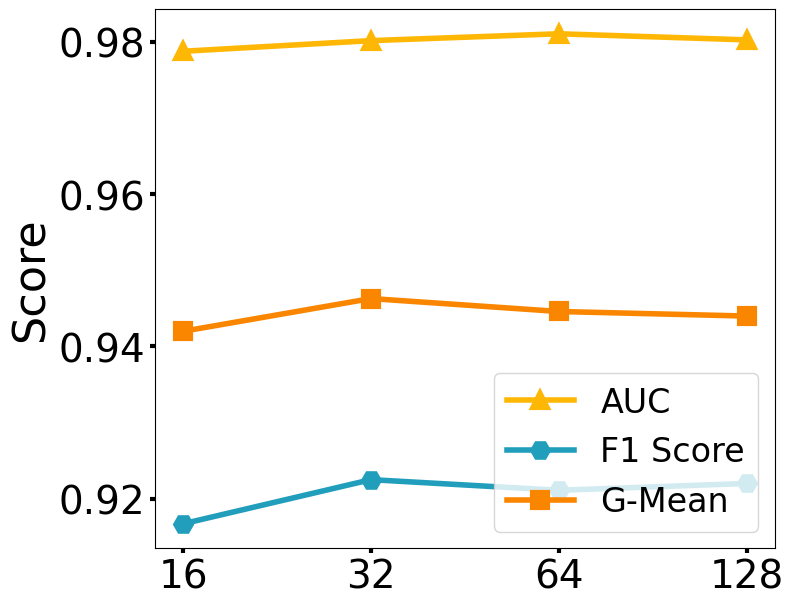}
    \caption{Hidden dimension}
  \end{subfigure}
  \hfill
  \begin{subfigure}[b]{0.49\linewidth}
    \includegraphics[width=\textwidth]{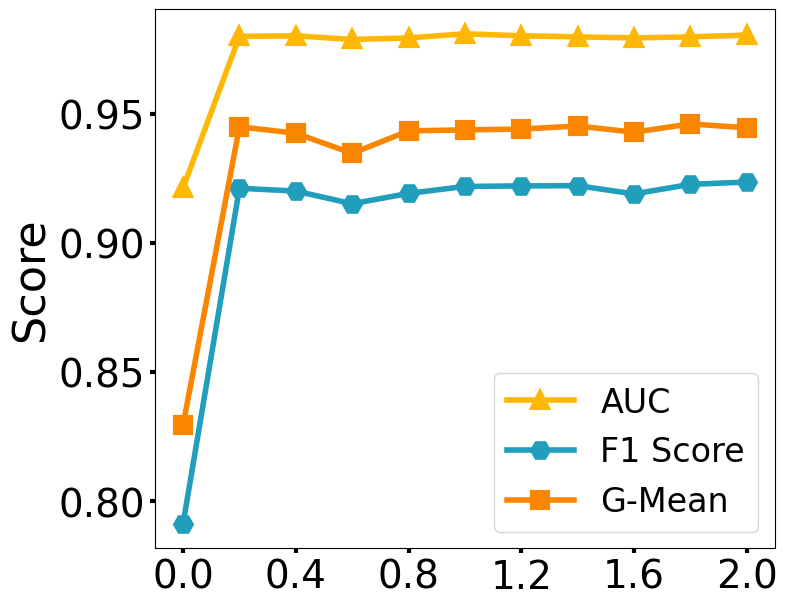}
    \caption{The weight $\gamma$}
  \end{subfigure}
  \begin{subfigure}[b]{0.49\linewidth}
    \includegraphics[width=\textwidth]{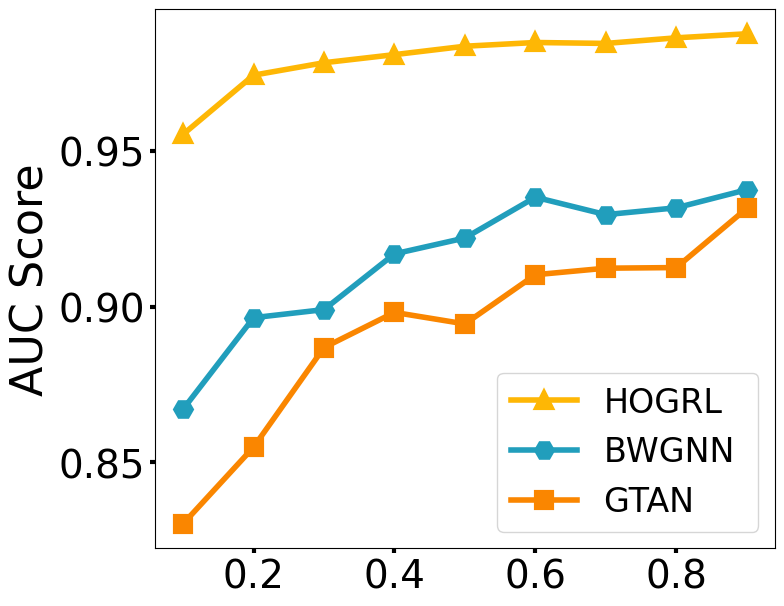}
    \caption{Train Ratio}
  \end{subfigure}
  \hfill
  \begin{subfigure}[b]{0.49\linewidth}
    \includegraphics[width=\textwidth]{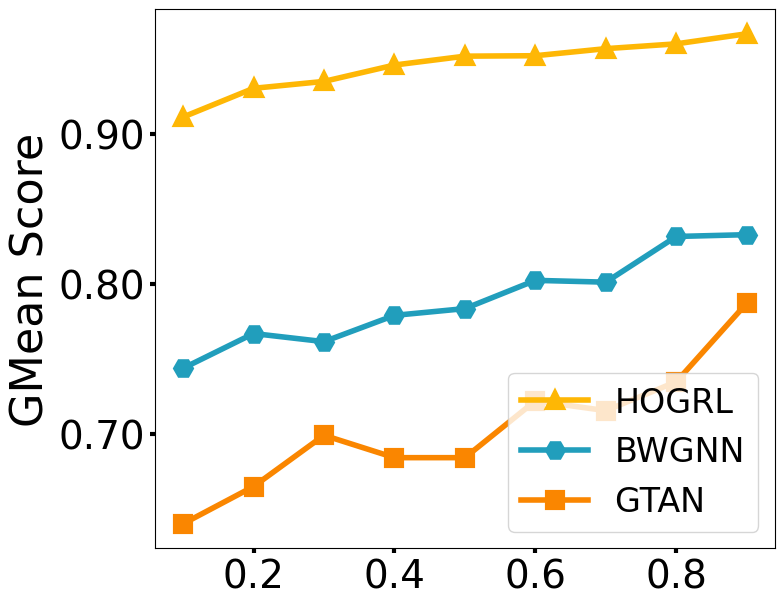}
    \caption{Train Ratio}
  \end{subfigure}
  \vspace{-5pt}
  \caption{Parameter sensitivity analysis with respect to (a) the hidden dimension; (b) the weight $\gamma$, (c) and (d) the train ratio.}
  \label{fig:parameters}
\end{figure}

We study the model parameter sensitivity by varying the hidden dimension, the weight $\gamma$ on the yelp dataset. Figure \ref{fig:parameters} (a) shows that when we
increase the hidden dimensions from 16 to 128, our model
maintains stable model performance and reaches the
performance peak at 64. As shown in fig\ref{fig:parameters} (b), We varied the weight hyperparameter $\gamma$ from 0 to 2. Performance reached its peak when $\gamma$=1. When the weight was 0, it corresponds to HOGRL/s. When the weight $\gamma$ varies between 0.2 and 2, the performance of our proposed HOGRL model shows no significant changes.

To further compare the learning capabilities of the models, we adjusted the proportion of the training set from 10\% to 90\%, with the remaining nodes equally divided between the validation set and the test set. For the sake of simplicity in our illustrations, we selected the state-of-the-art baseline model BWGNN and the superior semi-supervised learning model GTAN, and conducted experiments on the Yelp dataset. The results of these experiments are shown in Figure \ref{fig:parameters} (c) and (d). It can be observed that, regardless of the training ratio, HOGRL consistently performs well. Even with limited data (10\% training ratio), HOGRL still significantly outperforms both BWGNN and GTAN.
\subsection{Interpretability Exploration}
\begin{figure}
    \centering
  \begin{subfigure}[b]{0.49\linewidth}
    \includegraphics[width=\textwidth]{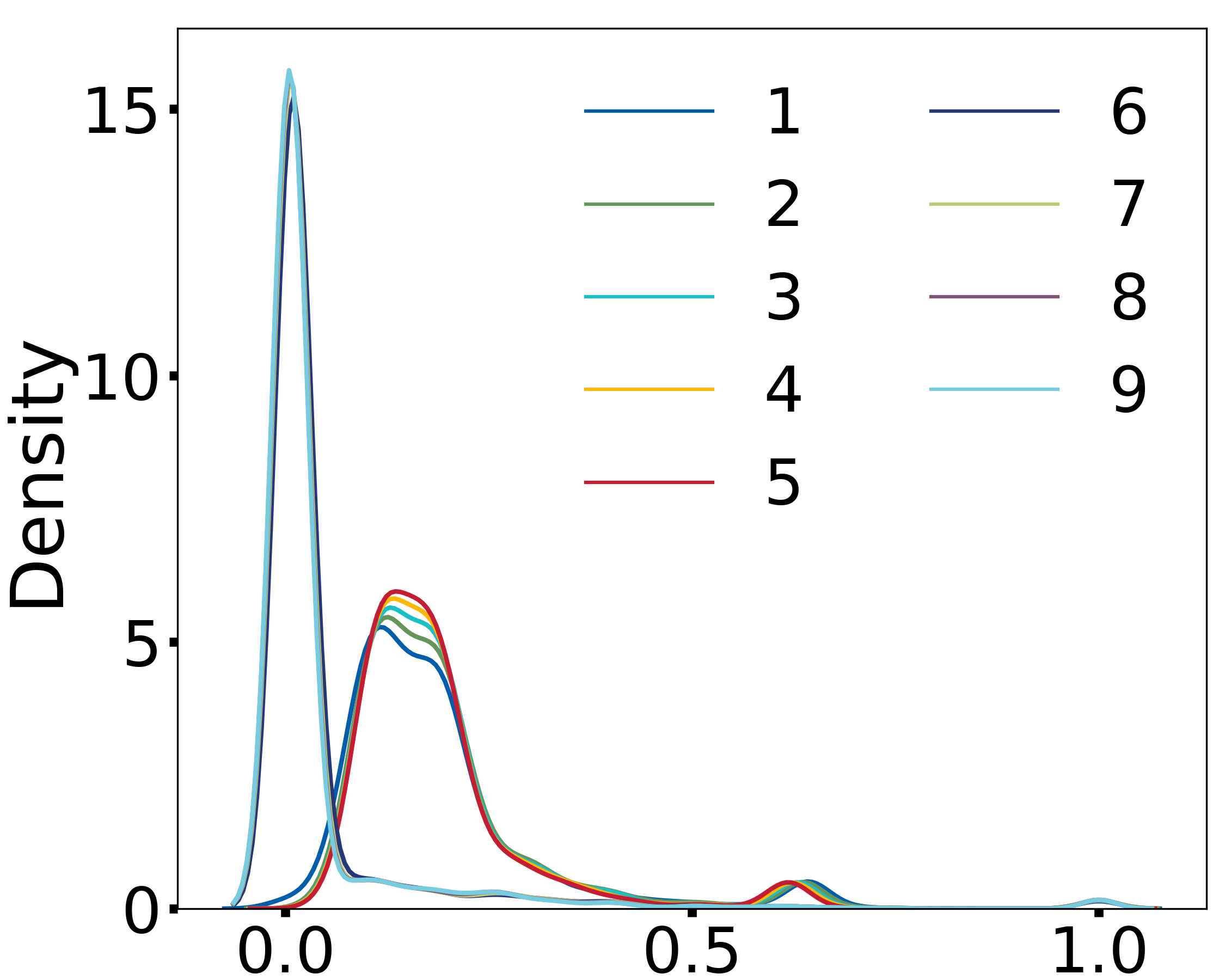}
    \caption{Traditional}
  \end{subfigure}
  \hfill
  \begin{subfigure}[b]{0.49\linewidth}
    \includegraphics[width=\textwidth]{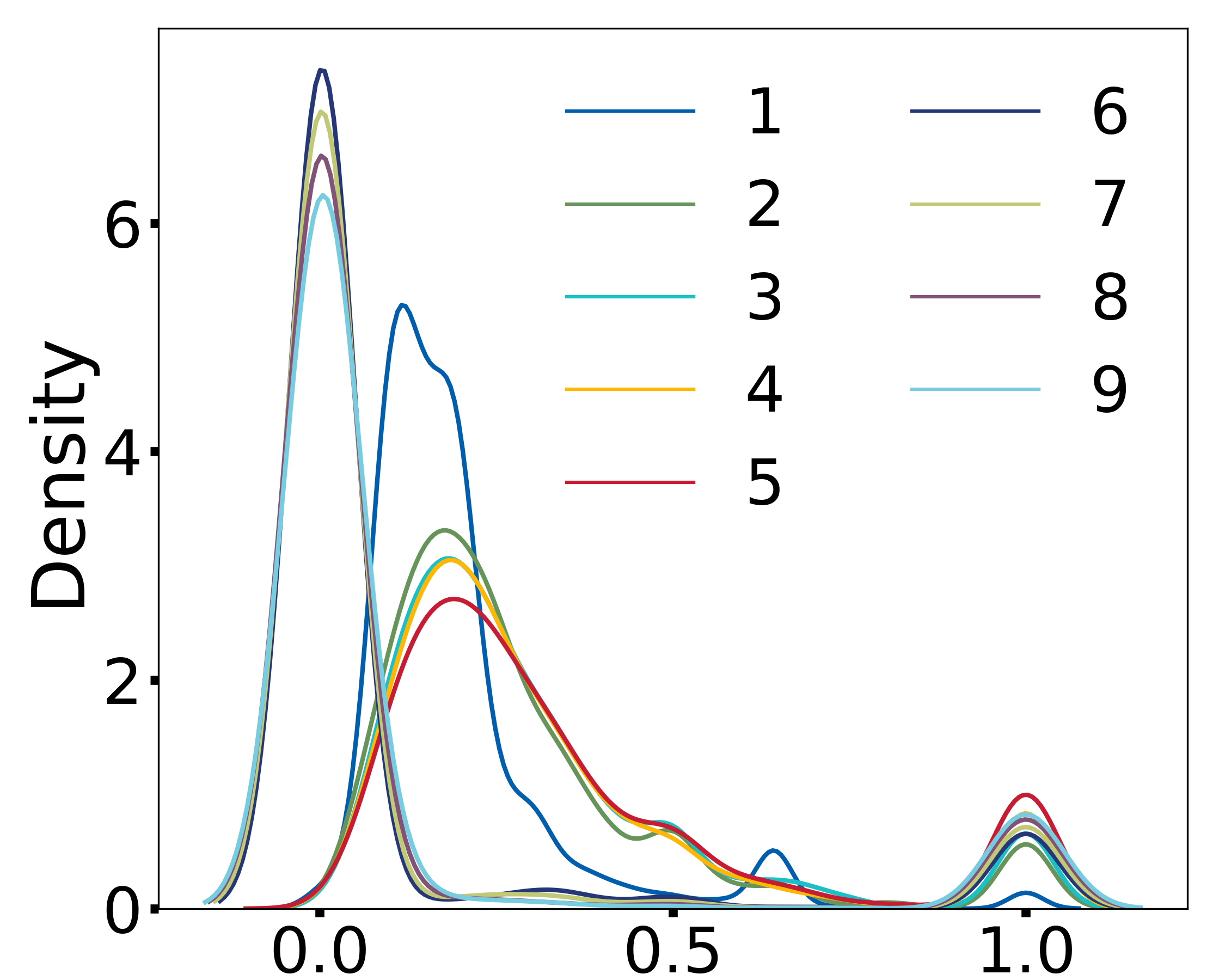}
    \caption{Ours}
  \end{subfigure}
    \caption{Homophily density distribution with different layers. The left side illustrates the homophily statistics of the traditional mixed-order propagation process, while the right side shows the homophily statistics of the high-order graphs proposed by HOGRL.}
    \label{fig:layers_homophily}
    \vspace{-8pt}
\end{figure}

We visualize the homophily density distribution of fraudulent nodes at different layers on yelp dataset in Figure \ref{fig:layers_homophily}. It can be observed that as the number of layers increases from 2 to 5, the peak density in the left figure remains nearly unchanged, whereas the peak density in the right figure shifts to the right. This indicates that the constructed high-order graph exhibits higher homophily. As the number of layers exceeds 6, the peak density on the left remains nearly constant, and the peak density on the left figure is at least twice that of the right figure. This demonstrates that the high-order graph constructed by HOGRL enhances the homophily of fraudulent nodes, reducing noise during the aggregation of high-order information. This is also the reason why HOGRL significantly outperforms HOGRL/s starting from L=6 layers.

\begin{figure}
  \centering
  \begin{subfigure}[b]{0.24\linewidth}
    \includegraphics[width=\textwidth]{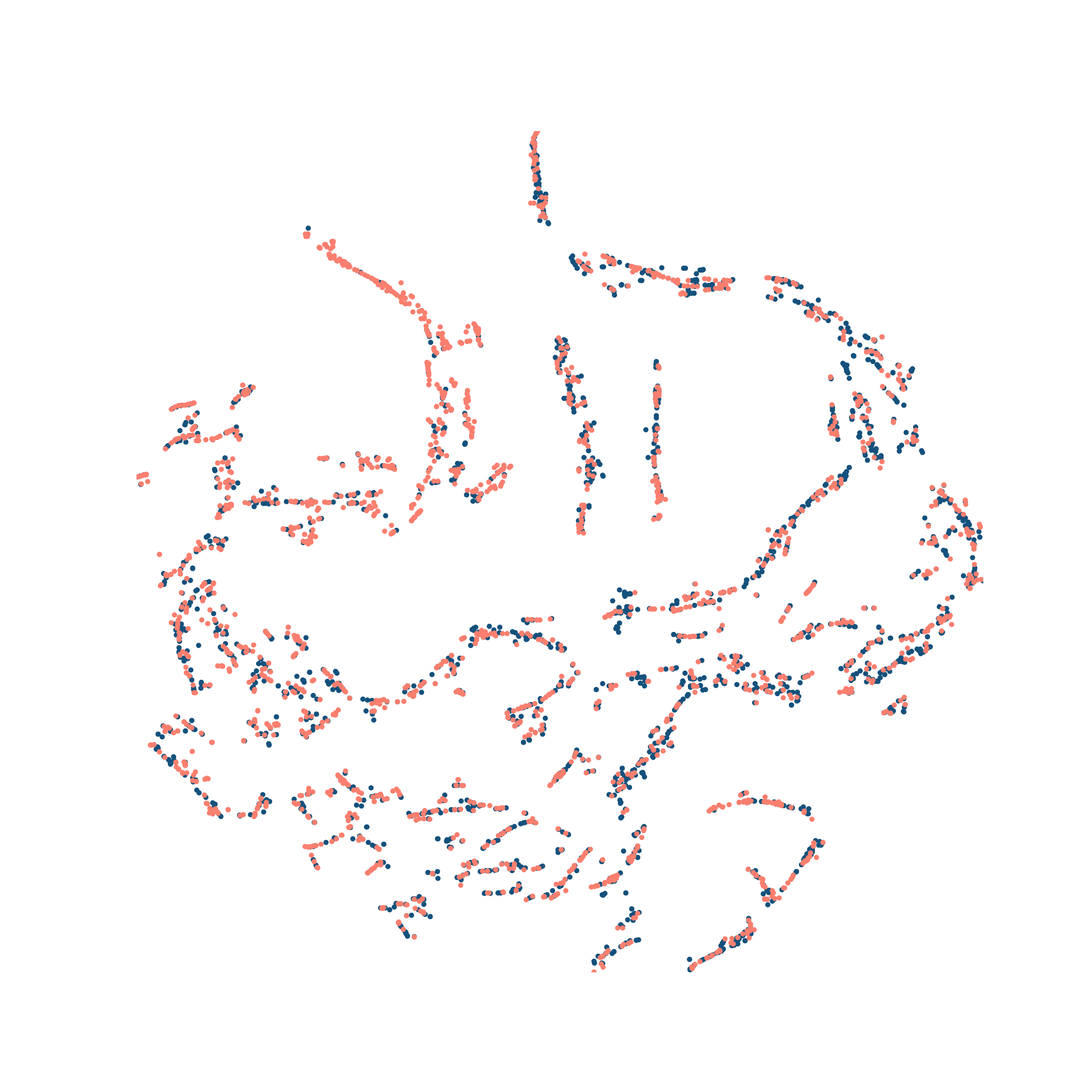}
    \caption{}
    \label{fig:gcn}
  \end{subfigure}
  \begin{subfigure}[b]{0.24\linewidth}
    \includegraphics[width=\textwidth]{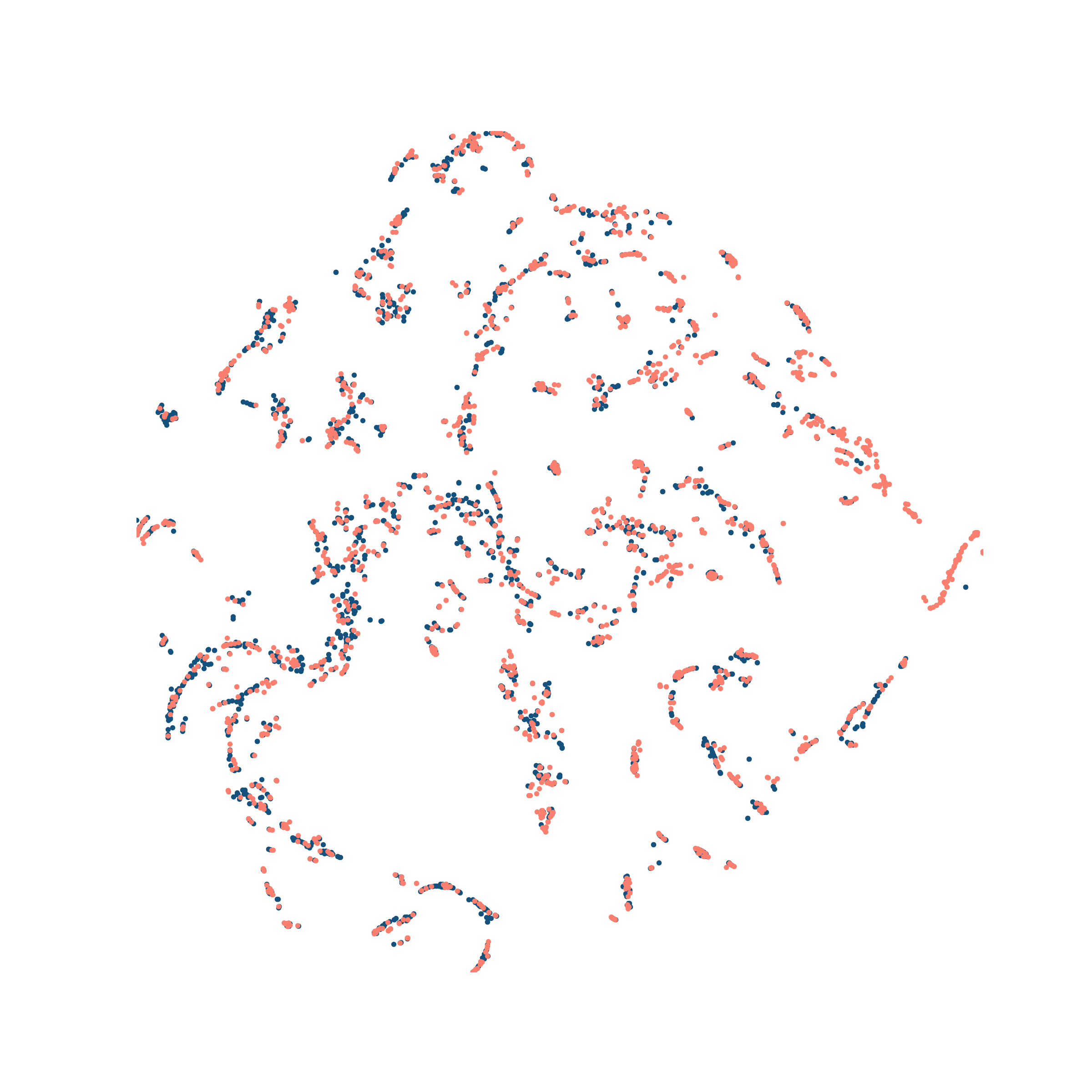}
    \caption{}
    \label{fig:gat}
  \end{subfigure}
  \begin{subfigure}[b]{0.24\linewidth}
    \includegraphics[width=\textwidth]{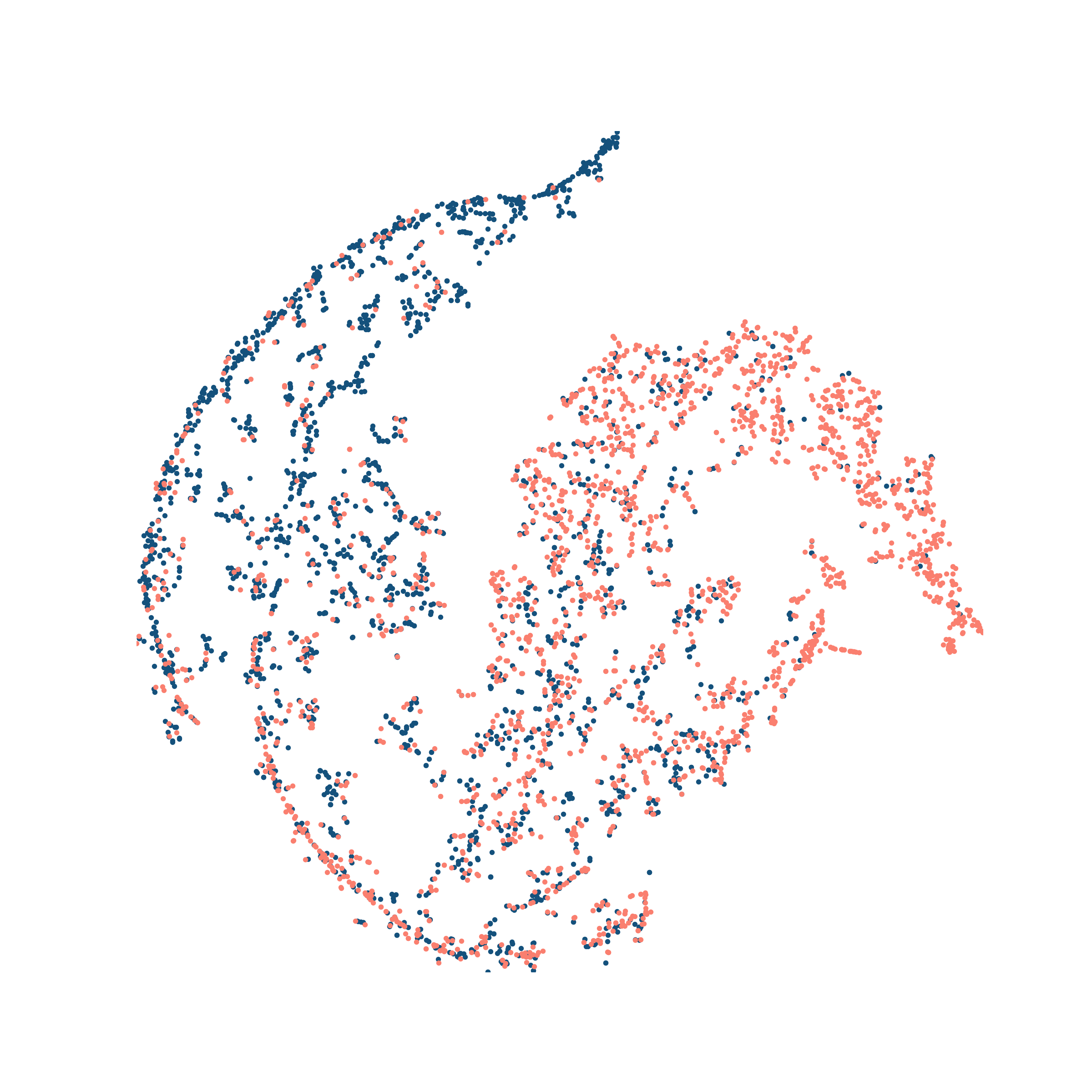}
    \caption{}
    \label{fig:graphsage}
  \end{subfigure}
  \begin{subfigure}[b]{0.24\linewidth}
    \includegraphics[width=\textwidth]{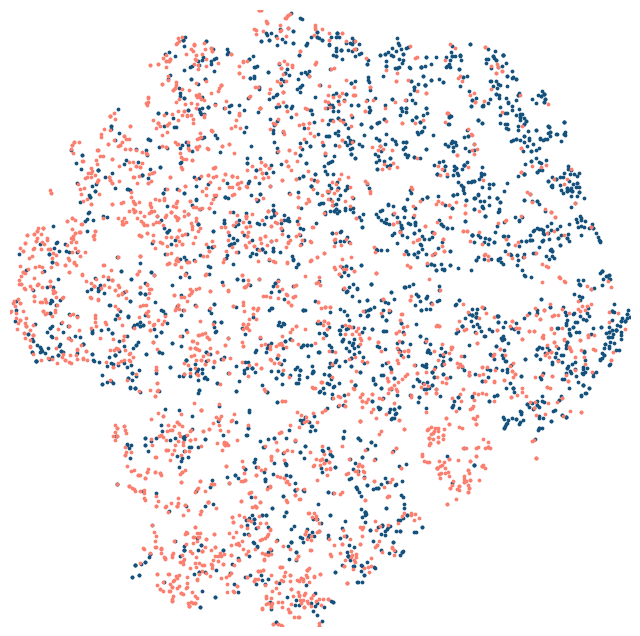}
    \caption{}
    \label{fig:fagcn}
  \end{subfigure}
  \begin{subfigure}[b]{0.24\linewidth}
    \includegraphics[width=\textwidth]{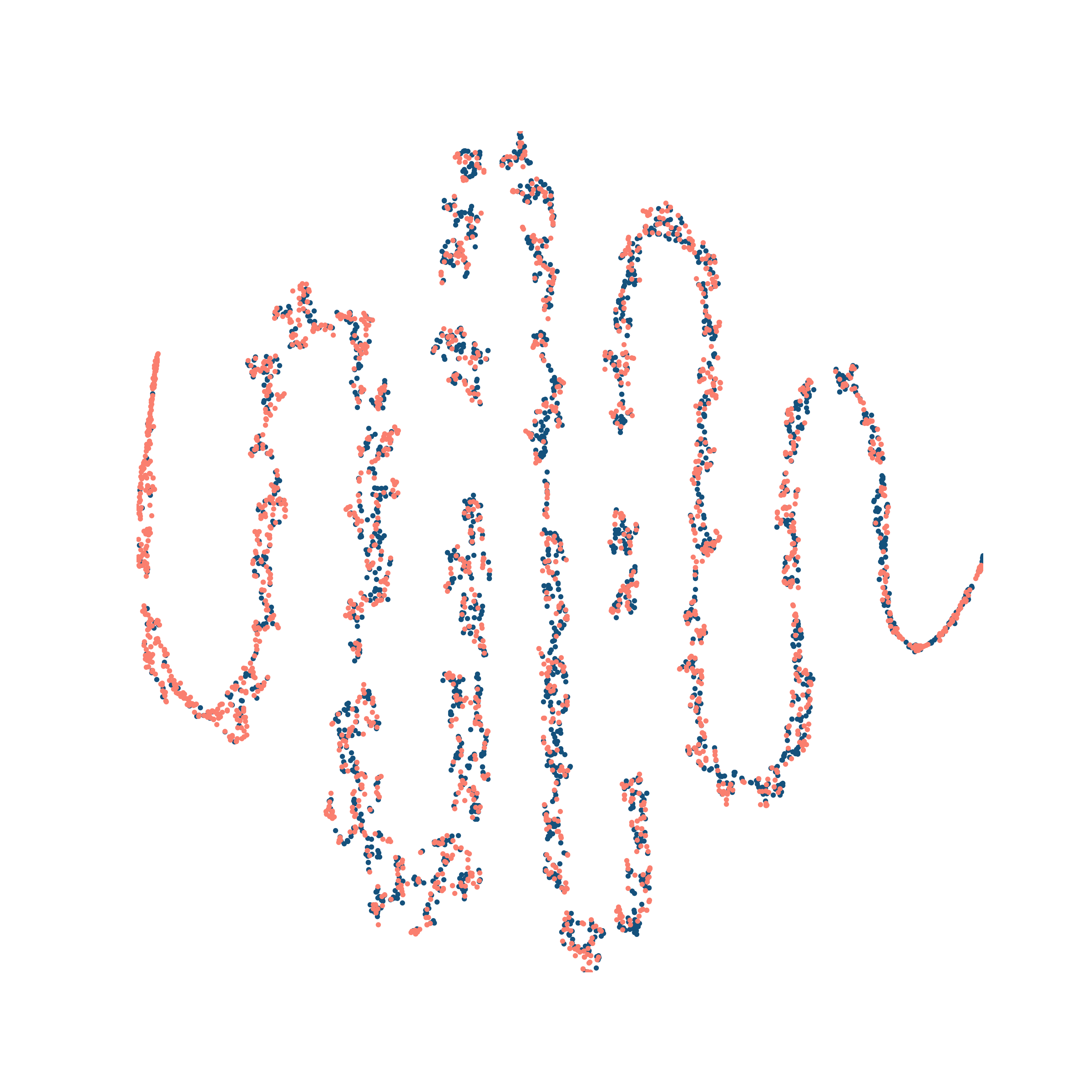}
    \caption{}
    \label{fig:GPRGNN}
  \end{subfigure}
  \begin{subfigure}[b]{0.24\linewidth}
    \includegraphics[width=\textwidth]{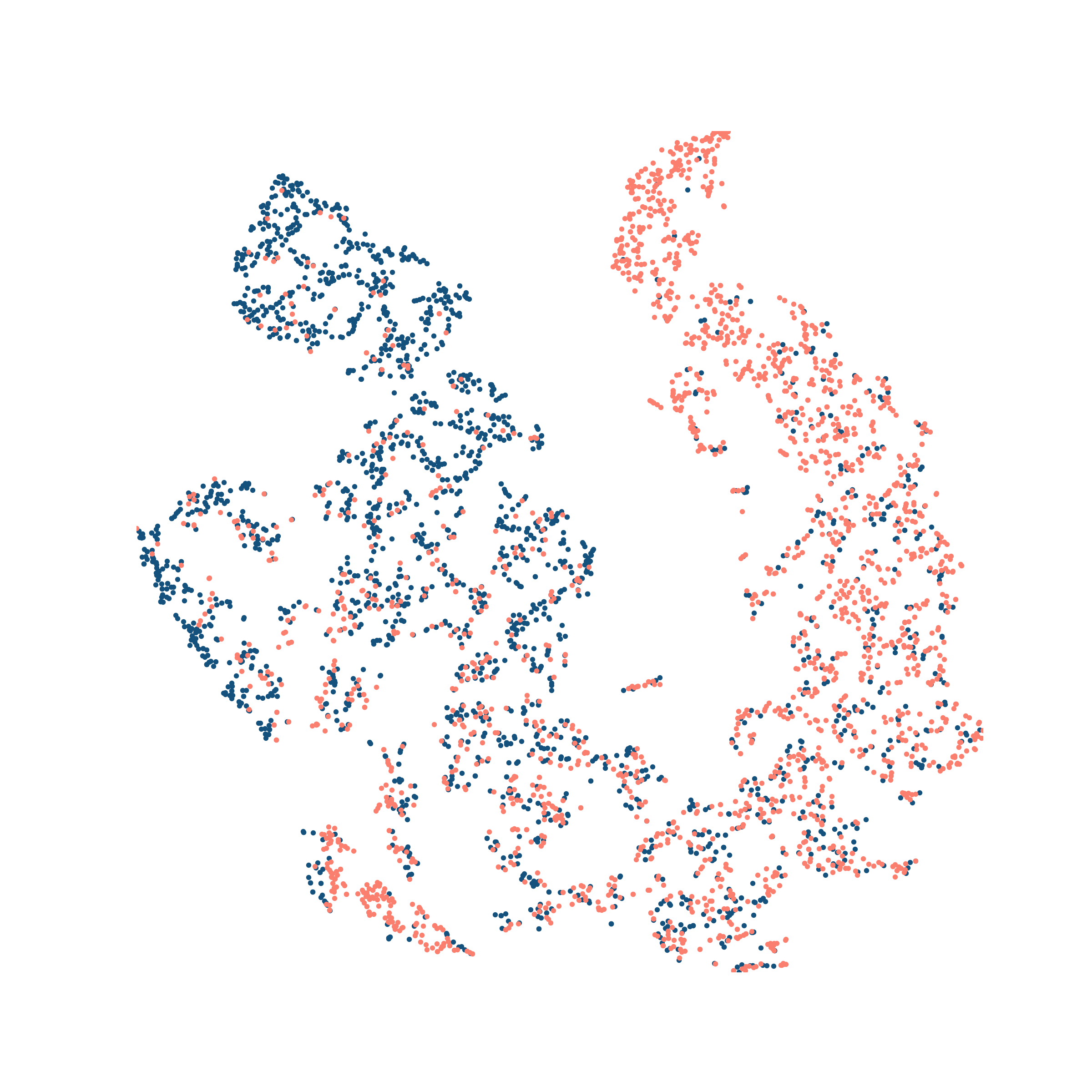}
    \caption{}
    \label{fig:GraphConsis}
  \end{subfigure}
  \begin{subfigure}[b]{0.24\linewidth}
    \includegraphics[width=\textwidth]{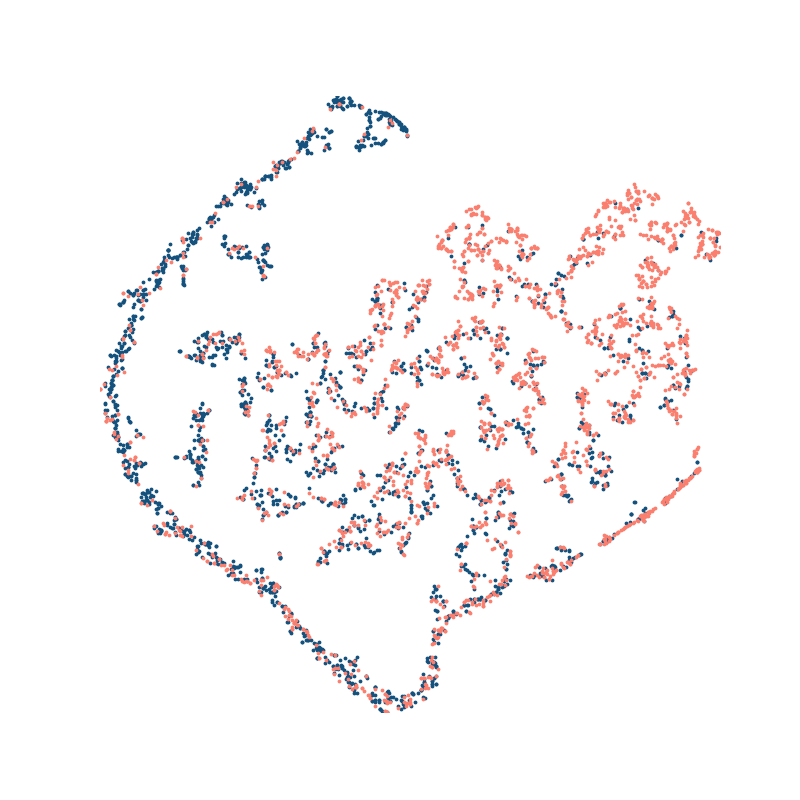}
    \caption{}
    \label{fig:CAREGNN}
  \end{subfigure}
  \begin{subfigure}[b]{0.24\linewidth}
    \includegraphics[width=\textwidth]{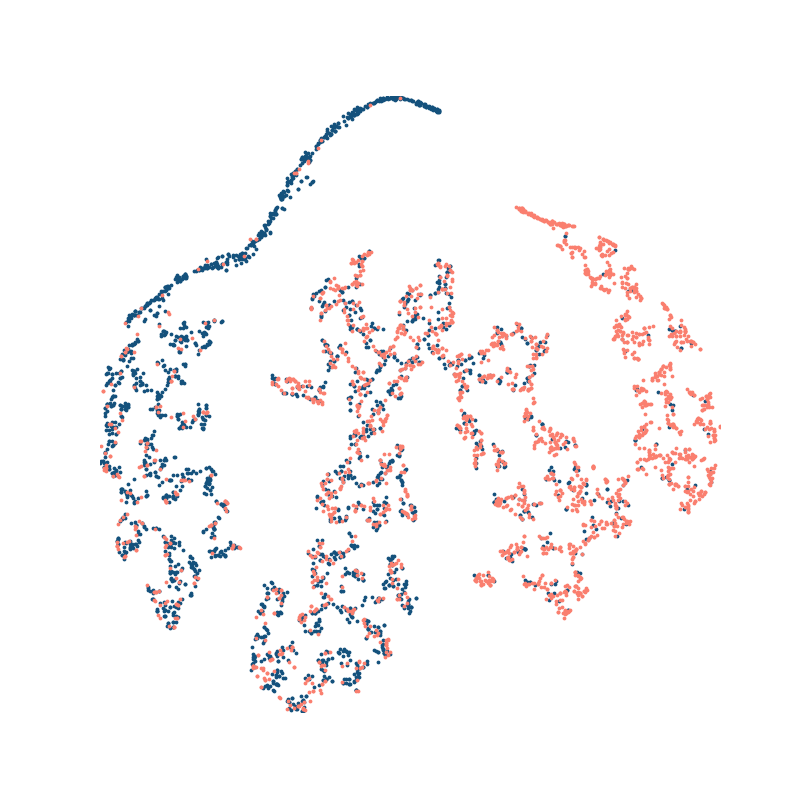}
    \caption{}
    \label{fig:PCGNN}
  \end{subfigure}
  \begin{subfigure}[b]{0.24\linewidth}
    \includegraphics[width=\textwidth]{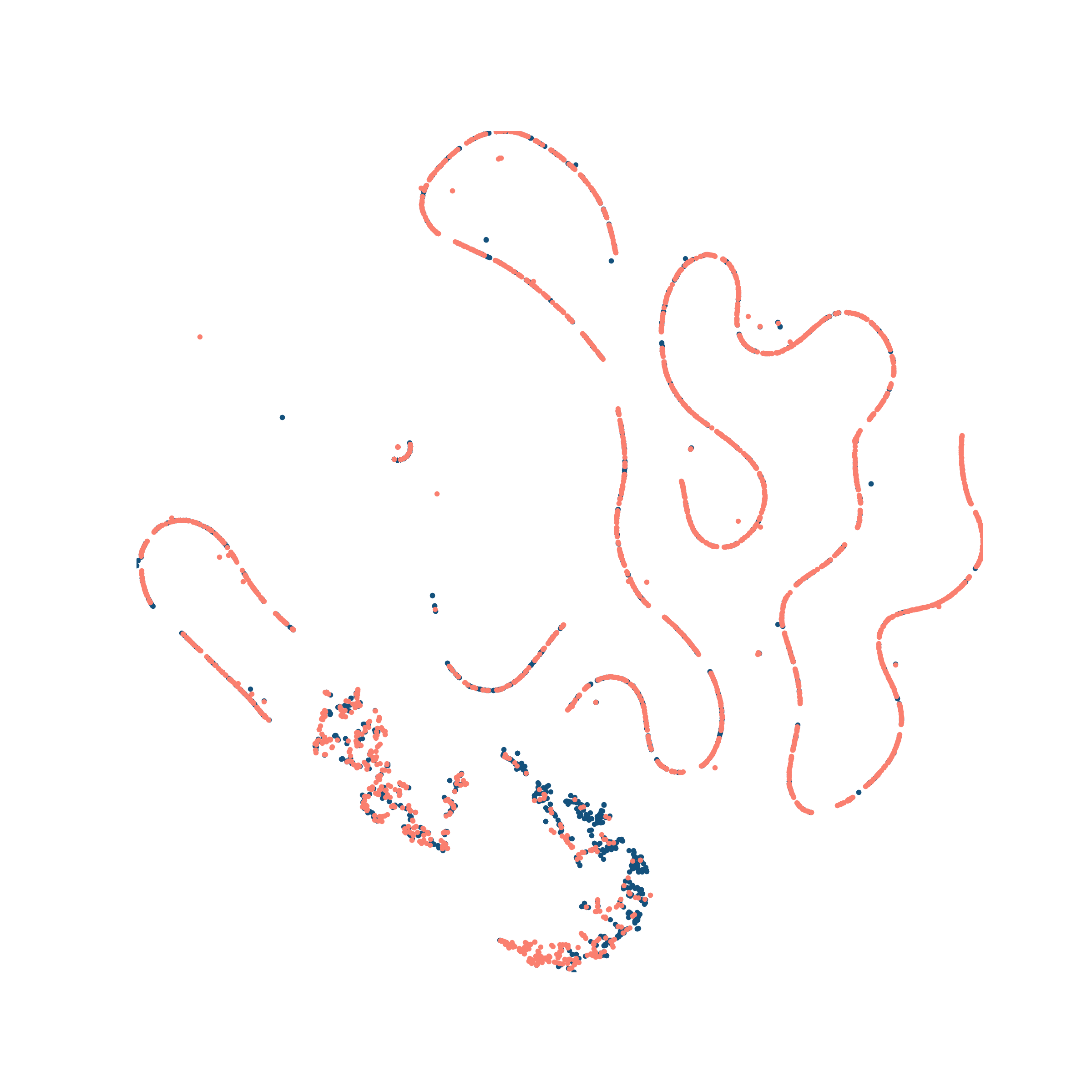}
    \caption{}
    \label{fig:H2}
  \end{subfigure}
  \begin{subfigure}[b]{0.24\linewidth}
    \includegraphics[width=\textwidth]{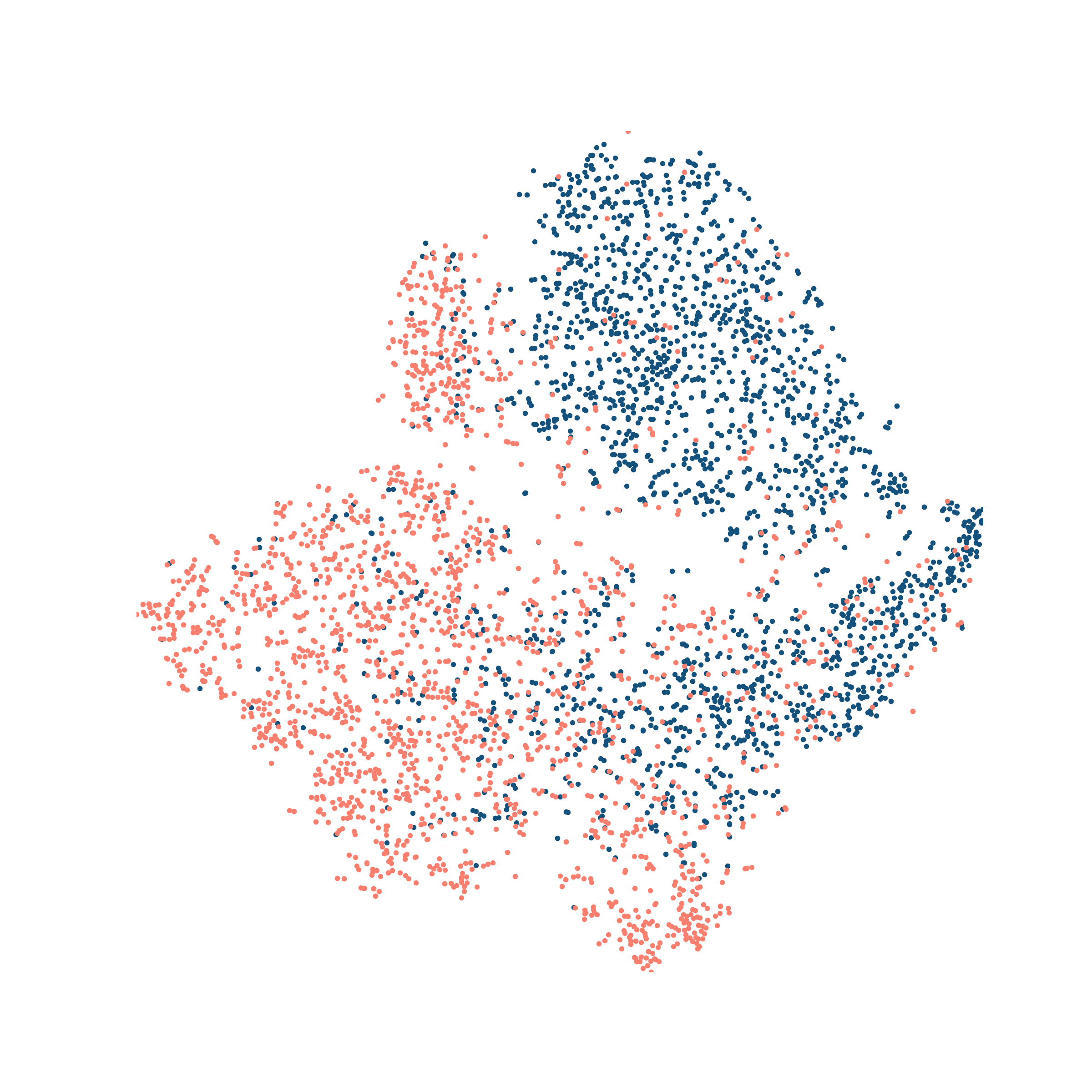}
    \caption{}
    \label{fig:GTAN}
  \end{subfigure}
  \begin{subfigure}[b]{0.24\linewidth}
    \includegraphics[width=\textwidth]{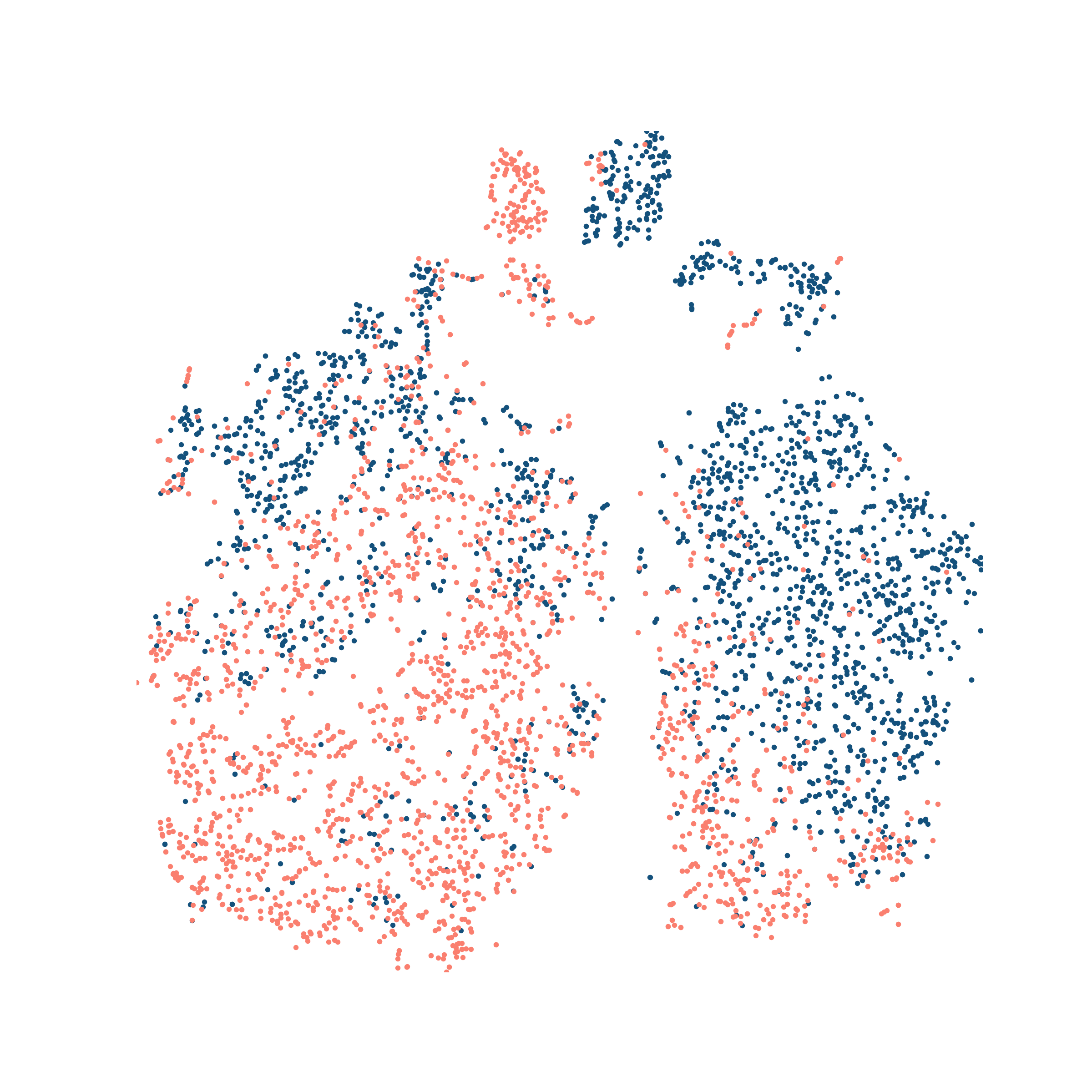}
    \caption{}
    \label{fig:BWGNN}
  \end{subfigure}
  \begin{subfigure}[b]{0.24\linewidth}
    \includegraphics[width=\textwidth]{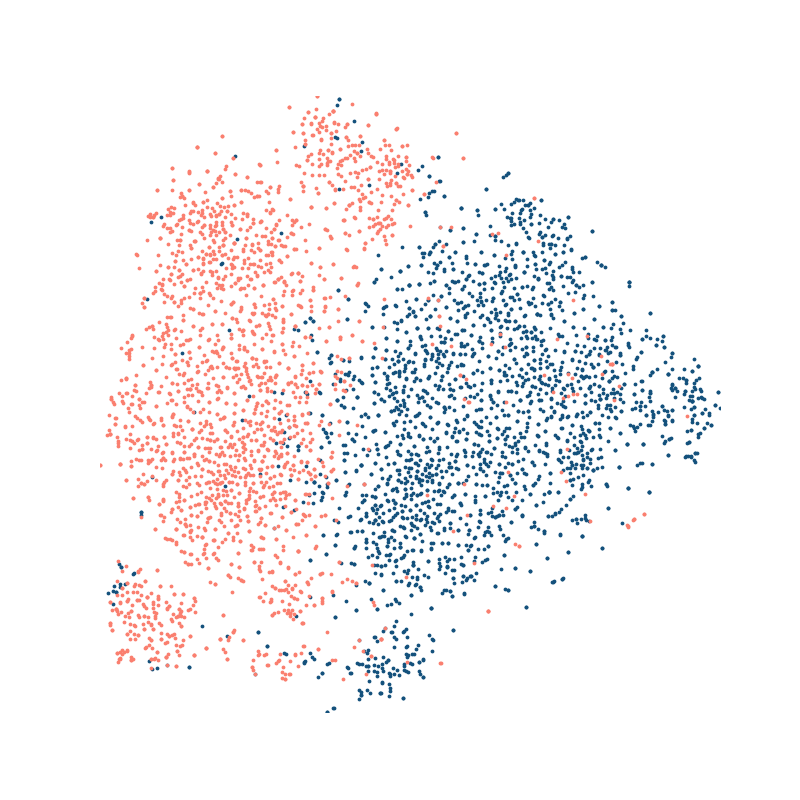}
    \caption{}
    \label{fig:HOGRL}
  \end{subfigure}
  \caption{Embedding visualization of different models. The red and blue nodes represent fraudsters and benign entities respectively.
  }
  \vspace{-8pt}
  \label{fig:visualization}
\end{figure}

We also conducted a visualization analysis of node embeddings on the Yelp dataset. To visually compare the performance of different models, we employed the t-SNE technique \cite{JMLR:v9:vandermaaten08a} to map the outputs from various models, just before their final layers, into a two-dimensional space for dimensionality reduction. This visualization method enabled us to clearly observe and analyze differences in the outputs of the models. The results, displayed in Figure \ref{fig:visualization}, show fraud nodes in red and benign nodes in blue, thus highlighting the distinction between the types of nodes. (a)-(k) represent the baseline models as listed in Table \ref{tab:2} in sequential order, and (l) depicts the visualization outcome for HOGRL. Compared to other models, the visualization of HOGRL distinctly shows a more effective separation between fraudulent and benign nodes. This improvement in discriminative ability is attributed to HOGRL directly learning distinct orders of pure representations from high-order transaction graphs, resulting in embeddings with greater distinction. For instance, compared to the BWGNN model, HOGRL demonstrates a notably reduced overlap between the two types of nodes.
\section{Conclusion}
In this paper, we propose a novel high-order graph representation learning model to reduce noise during multi-layer aggregation and identify disguised fraudsters involved in multi-hop indirect transactions. Specifically, we construct high-order transaction graphs and directly learn pure representations from them. Additionally, we introduce a mixture-of-expert attention mechanism to determine the significance of different orders. The comprehensive
experiments demonstrated the superiority of HOGRL compared with other baselines. The outstanding performance of HOGRL demonstrates its effectiveness in addressing high-order fraud camouflage crimes, maintaining the stability of the financial system, and positively influencing economic growth.
\section*{Acknowledgements}
This work was supported by the National Key R\&D Program of China (Grant no. 2022YFB4501704), the National Natural Science Foundation of China (Grant no. 62102287), the foundation of Key Laboratory of Artificial Intelligence, Ministry of Education, P.R. China and the Shanghai Science and Technology Innovation Action Plan Project (Grant no. 22YS1400600 and 22511100700).
\bibliographystyle{named}
\bibliography{ijcai24}

\end{document}